\pdfoutput=1  % arXiv: force pdflatex
\documentclass{article} % For LaTeX2e
\usepackage{iclr2027_conference,times}

\usepackage{amsmath,amsfonts,bm}

\def\eqref#1{equation~\ref{#1}}
\def\1{\bm{1}}

\DeclareMathAlphabet{\mathsfit}{\encodingdefault}{\sfdefault}{m}{sl}
\SetMathAlphabet{\mathsfit}{bold}{\encodingdefault}{\sfdefault}{bx}{n}

\usepackage[hidelinks]{hyperref}
\hypersetup{pdftitle={The Generator Is the Tracker: Multi-Object Tracking by Painting Persistent Identity Colours}, pdfauthor={Haiyu Yang, Miel Hostens}}
\usepackage{url}
\usepackage{booktabs}
\usepackage{multirow}
\usepackage{graphicx}
\graphicspath{{figures/}}
\usepackage{amsmath}
\usepackage{amssymb}
\usepackage{xcolor}
\newcommand{\rev}[1]{#1}

\title{The Generator Is the Tracker: Multi-Object Tracking\\by Painting Persistent Identity Colours}

\author{Haiyu Yang \quad Miel Hostens \\
Cornell University \\
\texttt{hy625@cornell.edu}
}

\iclrfinalcopy  % preprint build: authors shown, header reads 'Preprint'
\begin{document}

\maketitle

\begin{abstract}
Multi-object tracking (MOT) is conventionally decomposed into detection followed by
association, with object identity maintained as external state: track buffers, motion
models, appearance embeddings. We ask whether a video \emph{generator} can maintain that
state \emph{in pixels}. We fine-tune a 22B text-to-video diffusion model (LTX-2.3) with a
lightweight in-context LoRA to translate an RGB clip into an \emph{ID-map} clip, a video in
which every person is painted a flat, distinct colour that persists over time (same
colour, same identity). Long videos are generated as chained windows, each conditioned on
the cleaned tail of the previous one; a brief \emph{continuation} fine-tune teaches the
model to extend a given colouring, after which identity flows through the chain with no
separately trained detector, no motion model, and no association module. On the DanceTrack
test server the system reaches \textbf{40.3 HOTA}, to our knowledge the first entry whose
tracks are read directly off video produced by a pretrained video generator. This is well
below today's specialists (66--76 HOTA), but with an error profile no baseline shares: its
association score (\textbf{AssA 44.1}) exceeds every tracker of the 2020--22 benchmark
suite while detection is the deficit. \rev{A $2\times2$ factorial} on the
validation split shows where the identity comes from: \rev{neither chaining nor continuation
training helps on its own (16.8 and 15.9 HOTA against 18.2 for independent windows linked
post hoc), but together they reach 34.8}; re-linking the chain's own windows post hoc changes
nothing (34.7), and motion- and appearance-based associators run on the system's own boxes
reach only 22--24 AssA \rev{against 38.3 for the colours}, an ordering that holds \rev{in three sampling
seeds and} again on SportsMOT validation. On 383 mined occlusion \rev{and exit} events the chained
generator re-acquires 42\% of the identities it held on both sides of the gap, including
gaps longer than its 49-frame context\rev{; the pre-gap colour is painted during the gap in
every recovery but also in 110 of 119 failures, so carry-over is necessary but not
sufficient. We offer the system as a proof of concept, not a competitive tracker}. We
release code and checkpoints.
\end{abstract}

\begin{figure}[t]
\centering
\includegraphics[width=\linewidth]{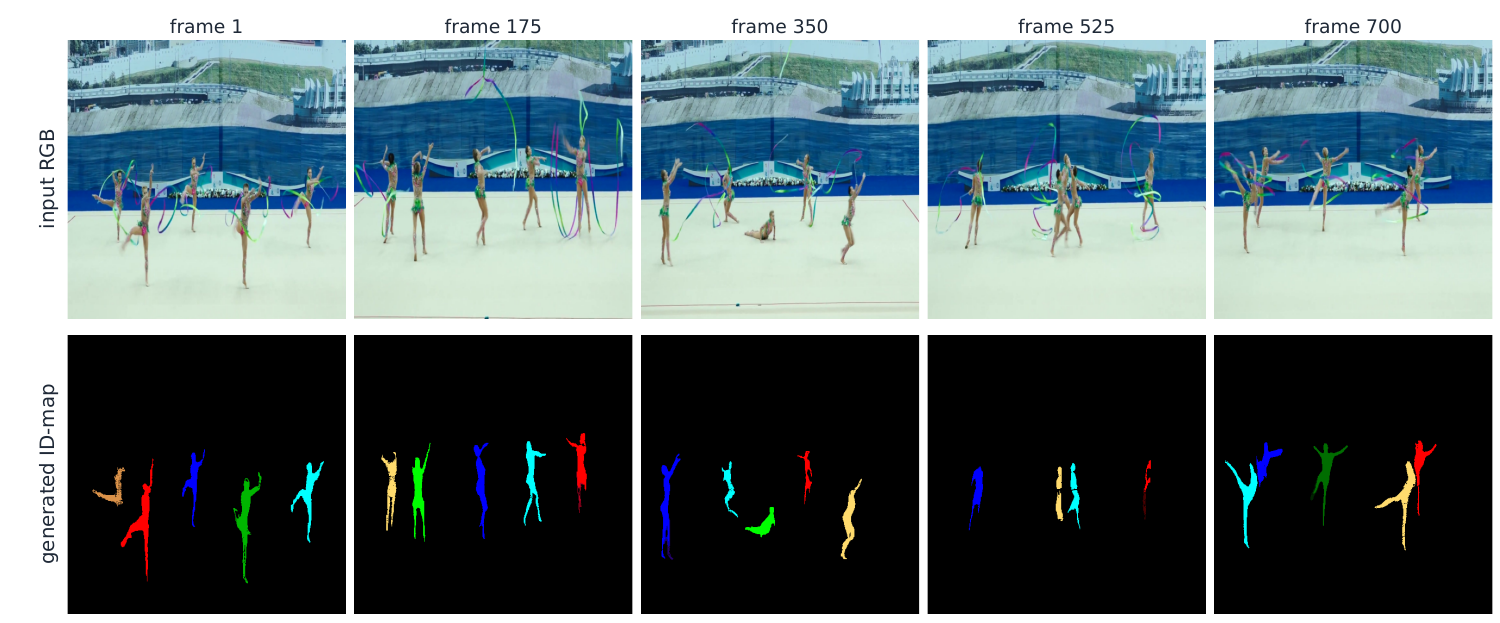}
\caption{\textbf{Tracking by painting.} Input RGB (top) and our generated ID-map video
(bottom) for a full 702-frame DanceTrack validation sequence (dancetrack0065, our
development video). Identity is carried entirely by the generated colours across 18 chained
generation windows. The five principal colours persist over the sequence (one dancer is
missed at frame 525); over the full 702 frames the chain scores IDF1 0.47 with 30 predicted
track ids for 5 people, i.e.\ persistence with short-lived fragments, not perfect identity.}
\label{fig:teaser}
\end{figure}

\section{Introduction}

The dominant paradigm in multi-object tracking is \emph{tracking-by-detection}: a detector
proposes boxes per frame, and an association stage stitches them into tracks using motion
models and appearance embeddings \citep{bewley2016sort,zhang2022bytetrack}. Even end-to-end
transformer trackers \citep{zeng2022motr} retain the same ontology: identity is
\emph{external state}, carried in query slots or track buffers alongside the video.

Video generative models suggest a different ontology. To synthesize a coherent video, a
generator must implicitly solve the binding problem: it must know which pixels belong to
the same object across frames, through occlusion, deformation, and camera motion. Such
correspondence is known to be present in diffusion features
\citep{tang2023dift,nam2025difftrack}. We ask whether it can be read out \emph{as
generation}: we ask a video diffusion model to emit, for an input RGB clip, a video of
\emph{ID-maps}, every tracked person painted a flat, distinct colour on black, with the
constraint that \emph{the same person keeps the same colour over time}
(Figure~\ref{fig:teaser}). Identity is not a slot, an embedding, or a learned buffer; it is
a property of the generated pixels, and decoding tracks requires only nearest-palette
colour quantisation. Our contributions are the mechanism and its controlled evaluation:

\begin{itemize}
\item \textbf{Identity-as-colour video translation.} We formalise tracking as RGB$\to$ID-map
video translation over a fixed $K{=}48$ CIELAB-separated palette with a lossless codec
(round-trip mask IoU $\geq 0.99$), and fine-tune LTX-2.3 with a rank-32 in-context LoRA
(positionally aligned reference conditioning) on 40 DanceTrack training videos with
silhouette supervision from SAM~3.

\item \textbf{Continuation training for unbounded video length.} Windows of 49 frames are
chained by conditioning each window's first 9 frames on the \emph{cleaned} (decoded and
re-encoded) tail of the previous window. Overlap-based chaining suffices for per-frame
targets such as depth \citep{hu2025depthcrafter}; for identity it fails at inference time,
because the model fights an imposed colouring it was never trained to continue. A brief
teacher-forced \emph{prefix} fine-tune teaches ``continue this colouring'' and raises
full-video IDF1 on the development video from 0.17 to 0.47.

\item \textbf{A generative tracker on the official evaluation server.} A single frozen
submission achieves \textbf{HOTA 40.3 / AssA 44.1 / IDF1 45.2}. We do not claim state of
the art: 2025--26 specialists reach 66--76 HOTA (Table~\ref{tab:test}). The result is the
error profile: association above every entry of the 2020--22 benchmark suite (maximum
40.2, MOTR) while detection (DetA 37.6) is the deficit, on the benchmark built to break
association; Section~\ref{sec:main} states what this comparison does and does not show.

\item \textbf{Where the identity comes from.} On the 25 full-length validation videos with
identical window geometry, \rev{a $2\times2$ factorial over chaining and continuation training
shows that neither helps alone: chaining without continuation training scores 16.8 HOTA and
continuation training without chaining 15.9, against 18.2 with neither (unchained windows
are linked post hoc); together they reach 34.8 (+18.9 [15.2, 22.9] over continuation
training alone).} Re-linking the chain's own windows with
the same associator changes nothing (34.7), and classical associators run on the chain's
own boxes reach AssA 21.6 (motion) and 24.3 (appearance and motion) against 38.3 for the
colours\rev{, a lead that holds in three sampling seeds}. Generation-time state \rev{that the
model is trained to continue}, not post-hoc association, carries identity.

\item \textbf{Re-acquisition through and beyond the context window.} On 383 mined
occlusion and exit events, the chained generator re-binds the correct identity for 42\% of
the 205 events it tracked on both sides of the gap (22\% of all 383); a Kalman/IoU
associator and an appearance-bank re-identifier on the same boxes and events reach 23\%
and 24\%. Gaps longer than the 49-frame context still re-acquire at 27\% (7 of 26).
\rev{Pre-registered statistics show what carries the identity: the pre-gap colour is painted
during the gap in every recovery, but also in 110 of 119 failures, so carry-over in pixels
is necessary but not sufficient; recoveries tend to keep the colour closer to the hidden
person's location (median IoU 0.35 against 0.19).}
\end{itemize}

\rev{We present the system as a proof of concept. Its absolute accuracy (40.3 HOTA) and
throughput (about 0.4 fps) are limitations, not claims. We report one failure as a
finding: crowded scenes fail even on training clips, where neighbours, mostly narrower than
one latent token, are often merged, and we cannot yet say whether resolution, capacity or
optimisation is the cause (Section~\ref{sec:ablations}).}

\section{Related Work}

\textbf{Multi-object tracking.} Tracking-by-detection pairs a detector with association by
motion \citep{bewley2016sort}, appearance \citep{wojke2017deepsort}, or both at scale
\citep{zhang2022bytetrack,zhang2021fairmot,zhou2020centertrack}. Transformer trackers make
association end-to-end but keep identity in query state
\citep{zeng2022motr,sun2020transtrack,gao2025motip,segu2025samba,shao2026fdta}\rev{, with
recent gains from better label assignment \citep{yan2025comot,yu2023motrv3} and learnable
association heads \citep{wang2025lamotr}}. Diffusion
has entered this stack as a box or motion denoiser inside tracking-by-detection
\citep{luo2024diffusiontrack,lv2024diffmot}; the generated object is a box, not the video.
TrackDiffusion \citep{li2025trackdiffusion} conditions video generation on tracklets; we
invert that mapping. DanceTrack \citep{sun2022dancetrack} was designed to defeat appearance
cues (uniform costumes, chaotic motion) and exposes association as the weak point of this
stack: strong detectors reach DetA 70--80 while AssA collapses to 20--40, which makes it
the natural testbed for a method whose only identity mechanism is the generator itself.

\textbf{Segmentation and foundation-model tracking.} SAM~2 \citep{ravi2025sam2} and SAM~3
\citep{carion2026sam3} track prompted objects with a per-object memory; SAM2MOT
\citep{jiang2026sam2mot} and SAMURAI++ \citep{yang2026samuraipp} turn this into
multi-object trackers by seeding SAM~2 from a detector and propagating one memory per
object, reaching 72--76 HOTA on DanceTrack without MOT training. These are the closest
neighbours to our system: identity also lives inside a foundation model rather than in a
tracker. They still need a box detector to instantiate objects and propagate objects one at
a time; we instantiate and bind all people jointly in one generated clip.

\textbf{Perception as generation.} Painter \citep{wang2023painter} and SegGPT
\citep{wang2023seggpt} cast perception as in-context image painting, SegGPT including
video object segmentation with masks painted as colours; Pix2Seq-D \citep{chen2023pix2seqd}
generates panoptic masks with a diffusion model and, in video, conditions on past
predictions so that instance ids persist frame to frame, the ancestor of our
decode-and-recondition chain\rev{. It is trained for panoptic segmentation and conditions
each frame on a few past predicted masks; we fine-tune a pretrained video generator
that paints a 49-frame window jointly}.
Tracking as colour propagation was observed in
self-supervised colourisation \citep{vondrick2018colorize}\rev{, where colour is an
\emph{input} signal: labels given in the first frame are copied forward by pointing, and
objects absent at $t{=}0$ are never instantiated. In our system colour is an \emph{output}
identity code: there are no first-frame masks, the model instantiates every person and
new entrant itself, and it must keep a colour alive while its person is invisible}.
Diffusion image generators have
been repurposed for dense prediction by light fine-tuning \citep{ke2024marigold}, and video
diffusion models for video depth with segment-wise inference and stitching
\citep{hu2025depthcrafter,chen2025videodepthanything}. Most directly,
\citet{gabeur2026imagegen} argue that image-generation pretraining plays the role LLM
pretraining plays for language, matching or surpassing specialists such as SAM~3 on several
segmentation benchmarks by instruction tuning an image generator, and name video input as
future work. Relative to this line we contribute (i) a pretrained 22B video generator
rather than a task model, (ii) whole-window generation with three-dimensional attention so
occlusions are bridged inside the context, (iii) explicit continuation training for
unbounded chaining, and (iv) evaluation as a tracker on an MOT test server.

\textbf{Correspondence in generative features.} DIFT \citep{tang2023dift} showed that
correspondence emerges in image diffusion features, and DiffTrack \citep{nam2025difftrack}
that video diffusion transformers encode temporal correspondence usable for zero-shot
point tracking. \rev{Both probe internal features for given query points; neither
instantiates objects, emits identities, or addresses identity through full occlusion.}
Our results extend this evidence to instance identity through full
occlusion, read out as generated pixels and measured on a tracking benchmark.

\textbf{Backbone.} We build on latent video diffusion \citep{rombach2022ldm,ho2022video}
and specifically LTX-2 \citep{hacohen2026ltx2,hacohen2024ltxvideo}, whose in-context LoRA
\citep{huang2024iclora} provides positionally aligned video-to-video conditioning and
whose timestep-based latent conditioning supports our prefix training. Flow matching
\citep{lipman2022flow} and LoRA \citep{hu2021lora} keep the fine-tune below 1\% of
parameters.

\section{Method}

\subsection{Identity as colour: the ID-map codec}
\label{sec:codec}

An \emph{ID-map} for frame $t$ paints instance $i$'s silhouette with palette colour
$c_{a(i)}$ on black, where $a$ is a per-video assignment into a fixed palette of $K{=}48$
colours chosen to maximise pairwise CIELAB separation (minimum $\Delta E = 26.4$). The
assignment is \emph{sequence-stable}: one colour per identity for the whole video. In
training data it maps a sequence's track ids to palette indices in sorted order; at
inference the generator chooses colours freely. Decoding inverts this by nearest-palette
quantisation in CIELAB followed by per-colour connected components with a minimum-area
filter (Appendix~\ref{app:params}); boxes are component bounding boxes and the palette
index is the track id. The codec
is frozen infrastructure: encode--decode round-trips preserve masks at IoU $\geq 0.99$ with
100\% identity agreement, and ID-maps are stored only in lossless formats (PNG, FFV1),
since lossy compression smears flat colours and silently corrupts decoding.

\subsection{Video-to-video fine-tuning with in-context LoRA}
\label{sec:iclora}

We fine-tune LTX-2.3 \citep{hacohen2026ltx2}, a 22B-parameter joint audio-video latent
diffusion transformer (48 dual-stream blocks; causal video VAE with $32\times$ spatial and
$8\times$ temporal compression), using its native \emph{in-context LoRA} mechanism: the
VAE-encoded reference video (RGB) is concatenated to the target token sequence with
\emph{aligned} positional encodings, kept clean (timestep 0), and excluded from the loss, so
bidirectional self-attention learns a positionally locked translation from reference to
target. We adapt only the video-stream projections (attention QKV and output, and the
feed-forward layers; rank 32; 164M trainable parameters, 0.74\% of the model) with a
flow-matching objective. Text conditioning is a fixed prompt whose embedding is
precomputed, which drops the 12B Gemma-3 text encoder from training memory: the entire
fine-tune fits on a single 80GB A100.

Training data are 777 clips of 49 frames at $512^2$ from the 40 DanceTrack training videos,
with silhouette supervision produced once by box-prompted SAM~3 \citep{carion2026sam3} on
ground-truth boxes (97\% of boxes matched back by IoU, filled-box fallback otherwise), and
occlusion-spanning clips oversampled. The validation split (25 videos) is never used for
training; it is used for checkpoint selection and method development
(Section~\ref{sec:experiments}).

\subsection{Continuation training: teaching the model to extend a colouring}
\label{sec:continuation}

A 49-frame window is far shorter than a video, and generating longer clips directly
degrades identity (Section~\ref{sec:ablations}); full videos therefore require
\emph{chained} windows with 9-frame overlaps (two latent frames). The chain conditions each
window on the previous one through the model's native prefix conditioning, which fixes the
first latent frames as clean tokens. The prefix is the previous window's cleaned tail: its
last 9 frames are decoded to palette colours, speckle-filtered, and re-encoded, so the
condition is always a valid ID-map rather than raw generation. We call this the decode,
correct, re-encode loop.

Applied at inference time only, this fails in an instructive way. The anchor binds the
overlap frames, but beyond them the model reverts to its own appearance-derived colours: it
was trained to \emph{invent} colourings, never to \emph{continue} one, and the errors
compound down the chain. We therefore resume training with an additional teacher-forced
prefix condition: with probability $0.5$, the target's own first two latent frames (9 pixel
frames) are given clean and excluded from the loss, teaching the skill ``propagate the
colouring you were handed.'' On our development video (dancetrack0065, a 5-person, 702-frame
validation sequence used for all single-video diagnostics) this single change lifts
full-video IDF1 from 0.172 to 0.466, while the $p{=}0.5$ mixture preserves single-window
mask quality for the chain's first window (foreground IoU 0.589 vs.\ 0.605) at a cost in
unconditioned identity quality (Section~\ref{sec:ablations}).

\begin{figure}[t]
\centering
\includegraphics[width=0.85\linewidth]{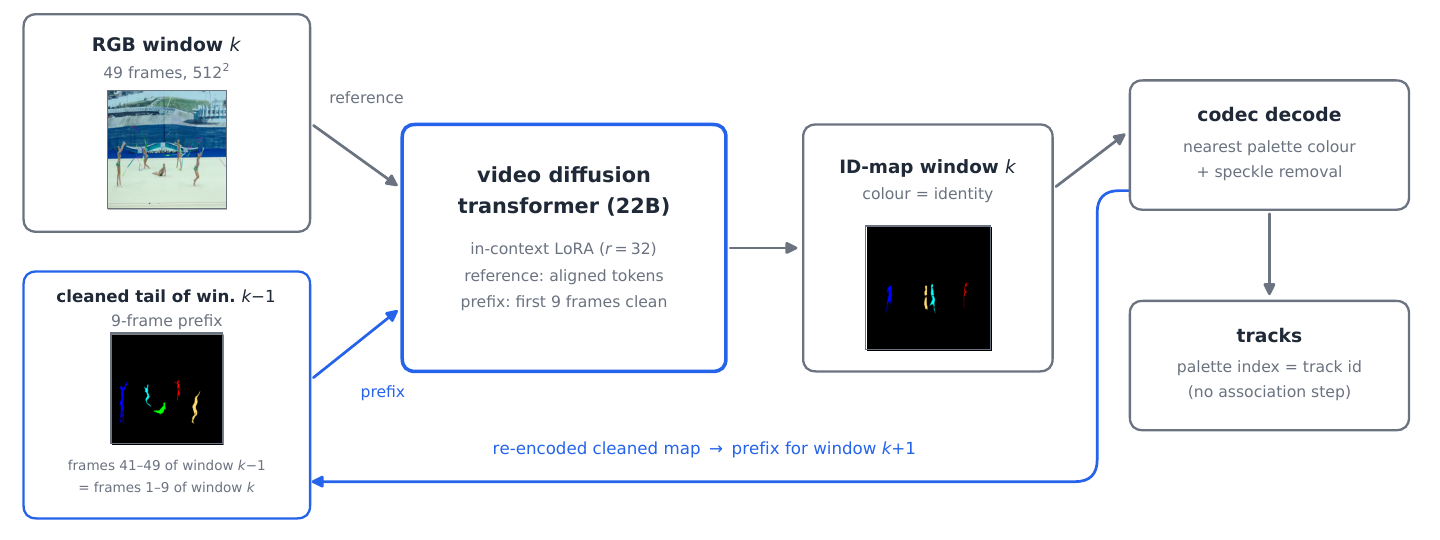}
\caption{\textbf{System overview.} Each 49-frame window is generated by a video diffusion
transformer conditioned on (i) the positionally aligned RGB reference (in-context LoRA) and
(ii) a 9-frame \emph{prefix}: the decoded, corrected (nearest palette colour, speckle
removal), and re-encoded tail of the previous window. The generated ID-map is decoded by
nearest-palette quantisation; the palette index \emph{is} the track id, and the cleaned map
seeds the next window's prefix.}
\label{fig:overview}
\end{figure}

\subsection{Inference: the full-video chain}
\label{sec:inference}

Figure~\ref{fig:overview} summarises the pipeline. Given a test video, we (i) split it into
49-frame windows at stride 40; (ii) generate window $k$ conditioned on [reference RGB$_k$;
prefix = cleaned tail of window $k{-}1$] with 20 denoising steps, classifier-free guidance
4.0 and spatio-temporal guidance 1.0, using the fixed prompt ``segmentation ID map: paint
each person a distinct flat colour on a black background''; (iii) decode, correct,
re-encode; (iv) emit boxes from the frames each window owns (overlaps split at their
midpoint), with same-colour components in a frame merged into one box. The colour index
\emph{is} the global track id: no separately trained detector and no association module
exist in the system; identity state lives in the generated pixels and is re-injected as
the prefix. New entrants receive whatever colour the model paints; nothing prevents the
reuse of a colour that belonged to someone who left, and such reuse surfaces as a merged
box. The palette bounds cumulative identities at $K{-}1=47$ per video
(Appendix~\ref{app:pervideo}).

\section{Experiments}
\label{sec:experiments}

\textbf{Setup.} Our primary benchmark is DanceTrack \citep{sun2022dancetrack}: 40 train /
25 val / 35 test videos (700--2,400 frames each) of group dance with near-identical costumes
and chaotic motion; Section~\ref{sec:sportsmot} repeats the recipe on SportsMOT
\citep{cui2023sportsmot}. Metrics are HOTA/DetA/AssA and their components \citep{luiten2021hota},
IDF1, and MOTA, computed with TrackEval (Appendix~\ref{app:params}). Test numbers come
from the official evaluation server (single submission, method frozen on validation, no
test-side iteration). The validation split is our model-selection and development set:
checkpoints were selected by validation foreground IoU, and the associator, decode
thresholds, and the continuation-training decision were developed on dancetrack0065, one of
the 25 videos in Tables~\ref{tab:ab}--\ref{tab:reacq}; the test row is the only
selection-free number. ``Pinned'' means fixed in a versioned log before the
corresponding run; any later change started a new labelled run. \rev{Main-table numbers
use sampling seed 0; two further seeds on all 25 validation videos are in
Section~\ref{sec:limitations} and Appendix~\ref{app:seeds}.} Total compute is about
\rev{224} A100-hours (Appendix~\ref{app:compute}).

\subsection{DanceTrack test-server results}
\label{sec:main}

\begin{table}[t]
\caption{DanceTrack \textbf{test} results (official server). Baseline values are from the
DanceTrack benchmark table as updated by the benchmark authors (arXiv v3 of
\citealp{sun2022dancetrack}, Table~3, which re-ran QDTrack with YOLOX and added OC-SORT
after publication; reproduced in \citealp{zhang2023motrv2}) and from each cited method
paper. Det.: sep.\ = separately trained box detector; e2e = detection queries trained
jointly; SAM~2 = detector-seeded SAM~2 memory. TrackTrack does not report DetA.
\rev{$^{+}$Uses CrowdHuman, in joint training (CO-MOT reports 65.3 HOTA without it) or for
its proposal detector (MOTRv2).
$^{\ddagger}$Training-time pseudo-labels from detectors that were themselves trained with
extra data, including CrowdHuman. Ours uses no tracking data beyond
the DanceTrack train split, but a web-pretrained generator and SAM~3 silhouettes, a
different kind of prior from detector pretraining.}}
\label{tab:test}
\begin{center}
\scriptsize
\setlength{\tabcolsep}{2.4pt}
\begin{tabular}{llcccccccc}
\toprule
 & Method & Year & Det. & HOTA$\uparrow$ & DetA$\uparrow$ & AssA$\uparrow$ & IDF1$\uparrow$ & MOTA$\uparrow$ \\
\midrule
\multirow{9}{*}{\rotatebox{90}{\scriptsize 2020--22 suite}}
& CenterTrack \citep{zhou2020centertrack} & 2020 & sep. & 41.8 & 78.1 & 22.6 & 35.7 & 86.8 \\
& TransTrack \citep{sun2020transtrack} & 2020 & e2e & 45.5 & 75.9 & 27.5 & 45.2 & 88.4 \\
& FairMOT \citep{zhang2021fairmot}     & 2021 & sep. & 39.7 & 66.7 & 23.8 & 40.8 & 82.2 \\
& TraDeS \citep{wu2021trades}          & 2021 & sep. & 43.3 & 74.5 & 25.4 & 41.2 & 86.2 \\
& QDTrack (YOLOX re-run)               & 2021 & sep. & 54.2 & 80.1 & 36.8 & 50.4 & 87.7 \\
& ByteTrack \citep{zhang2022bytetrack} & 2022 & sep. & 47.7 & 71.0 & 32.1 & 53.9 & 89.6 \\
& GTR \citep{zhou2022gtr}              & 2022 & sep. & 48.0 & 72.5 & 31.9 & 50.3 & 84.7 \\
& MOTR \citep{zeng2022motr}            & 2022 & e2e & 54.2 & 73.5 & 40.2 & 51.5 & 79.7 \\
& OC-SORT \citep{cao2023ocsort}        & 2023 & sep. & 55.1 & 80.3 & 38.3 & 54.6 & 92.0 \\
\midrule
\multirow{5}{*}{\rotatebox{90}{\scriptsize 2023--24}}
& Deep OC-SORT \citep{maggiolino2023deepocsort} & 2023 & sep. & 61.3 & 82.2 & 45.8 & 61.5 & 92.3 \\
& MeMOTR \citep{gao2023memotr}         & 2023 & e2e & 68.5 & 80.5 & 58.4 & 71.2 & 89.9 \\
& MOTRv2\rev{$^{+}$} \citep{zhang2023motrv2}       & 2023 & sep.+e2e & 69.9 & 83.0 & 59.0 & 71.7 & 91.9 \\
& \rev{MOTRv3 \citep{yu2023motrv3}}    & \rev{2023} & \rev{e2e$^{\ddagger}$} & \rev{70.4} & \rev{83.8} & \rev{59.3} & \rev{72.3} & \rev{92.9} \\
& DiffMOT \citep{lv2024diffmot}        & 2024 & sep. & 62.3 & 82.5 & 47.2 & 63.0 & 92.8 \\
\midrule
\multirow{7}{*}{\rotatebox{90}{\scriptsize 2025--26}}
& TrackTrack \citep{shim2025tracktrack} & 2025 & sep. & 66.5 & -- & 52.9 & 67.8 & 93.6 \\
& \rev{CO-MOT$^{+}$ \citep{yan2025comot}} & \rev{2025} & \rev{e2e} & \rev{69.4} & \rev{82.1} & \rev{58.9} & \rev{71.9} & \rev{91.2} \\
& MOTIP \citep{gao2025motip}           & 2025 & e2e & 69.6 & 80.4 & 60.4 & 74.7 & 90.6 \\
& \rev{LA-MOTR \citep{wang2025lamotr}} & \rev{2025} & \rev{e2e} & \rev{71.1} & \rev{84.4} & \rev{59.0} & \rev{71.8} & \rev{93.5} \\
& FDTA \citep{shao2026fdta}            & 2026 & e2e & 71.7 & 81.0 & 63.5 & 77.2 & 91.3 \\
& SAMURAI++ \citep{yang2026samuraipp}  & 2026 & sep.+SAM~2 & 72.8 & 82.2 & 64.7 & 77.7 & 92.1 \\
& SAM2MOT \citep{jiang2026sam2mot}     & 2026 & sep.+SAM~2 & 75.8 & 79.7 & 72.2 & 83.9 & 88.5 \\
\midrule
& \textbf{Ours (generation only)} & 2026 & none & 40.3 & 37.6 & \textbf{44.1} & 45.2 & 16.2 \\
\bottomrule
\end{tabular}
\end{center}
\end{table}

\begin{figure}[t]
\centering
\includegraphics[width=0.55\linewidth]{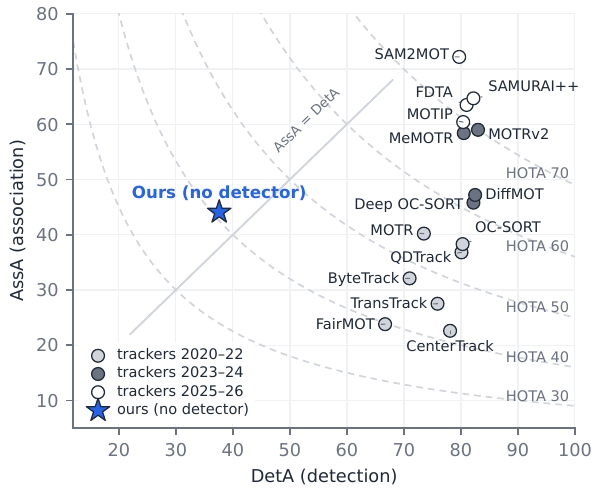}
\caption{\textbf{The generative tracker inverts the classical error profile.} DanceTrack
test decomposition: every detector-based tracker sits below the AssA${}={}$DetA diagonal
(detection ahead of association), while ours (star) is the only method above it and the
only one without a detector. Modern specialists (upper right) exceed our absolute scores on
both axes. Dashed curves are iso-HOTA. Values as in Table~\ref{tab:test}\rev{; MOTRv3,
CO-MOT and LA-MOTR lie beside MOTRv2, undrawn}.}
\label{fig:scatter}
\end{figure}

Table~\ref{tab:test} and Figure~\ref{fig:scatter} show the result. This is a paradigm
result, not a state-of-the-art result: 2025--26 specialists reach 66--76 HOTA, and our 40.3
sits with the 2020--21 detector-based systems. What no other entry offers is the profile.
Ours is, to our knowledge, the only submission that runs neither a box detector nor an
association or propagation module, and the only one whose association exceeds its
detection. AssA 44.1 is above every tracker of the 2020--22 suite, including the end-to-end
transformer MOTR (40.2) and the motion specialist OC-SORT (38.3 in the benchmark table;
40.2 in its own paper), while DetA (37.6 against 80 for specialists) is the entire deficit.
Modern systems reach higher association (\rev{45.8}--72.2) through stronger association
machinery: adaptive re-identification \citep{maggiolino2023deepocsort}, non-linear motion
models \citep{lv2024diffmot}, long-term query memory \citep{gao2023memotr}, detector
proposals \citep{zhang2023motrv2}, or a per-object foundation-model memory
\citep{jiang2026sam2mot}; their AssA/DetA ratio stays below one (SAM2MOT is closest at
0.91), whereas ours is 1.17.

\textbf{What the profile does and does not show.} AssA is computed only over matched
detections, so a tracker that detects the easy people and keeps their tracks whole scores
high AssA at low DetA. Our recall is low (validation DetRe 41.4 / DetPr 51.3; LocA 77.1, so
localisation of what is found is not the problem) and lowest in crowds, where association
also falls (Figure~\ref{fig:density}, Appendix~\ref{app:pervideo}). AssA above the 2020--22
suite is therefore a statement about association on the people the system finds, which
over-represents sparse scenes; the like-for-like test is association on \emph{identical}
detections, where classical associators on our own boxes fall well short of the colours
(Section~\ref{sec:ab}). Test scores exceed validation scores (34.8 HOTA / 38.3 IDF1) with a
single frozen submission; the 35 test videos are on average sparser than the 25 validation
videos, as the density dependence predicts.

\subsection{Where does the identity come from? Controlled comparisons}
\label{sec:ab}

Identity could, in principle, be recovered from generated windows by classical means\rev{:
generate every window independently and link the windows afterwards. That control differs
from the chain in two ways at once: the chain conditions each window on the previous one,
and it uses the checkpoint after continuation training (\emph{prefix-trained}), not the one
before it (\emph{translation-only}). We therefore ran a full $2\times2$ factorial on the 25
full-length validation videos with the same window geometry (640 windows per cell), decoder
and sampler (Table~\ref{tab:ab}): each checkpoint, run as independent windows or as a
chain. Cell A is the chain of Section~\ref{sec:inference} and cell B the control just
described (arms A and B in what follows); C chains the translation-only checkpoint; D
generates independent windows with the prefix-trained one. Independent windows share no state, so they are linked
post hoc by a pinned overlap-IoU associator (Appendix~\ref{app:stitcher}); the chained cells
read identity off the colour, and each cell is also scored under the other link. Inside
every cell we further drop the ids from that cell's own boxes and re-associate them with a
SORT-style Kalman/IoU tracker \citep{bewley2016sort} and with an appearance-and-motion
identity bank over DINOv3 embeddings (Appendix~\ref{app:params}), which compares
association on identical detections. Differences are paired per-video bootstrap means with
95\% intervals (10,000 resamples) and the number of videos favouring the first term.}

\begin{table}[t]
\caption{\rev{Checkpoint $\times$ generation factorial on identical window geometry: 25
full-length validation videos, TrackEval, sampling seed 0. Each cell is read under its
native link (colour for chains, the post-hoc associator for independent windows). The last
two columns re-associate that cell's own boxes, so association is compared at fixed
detection inside every cell. Paired effects are in the text.}}
\label{tab:ab}
\begin{center}
\scriptsize
\setlength{\tabcolsep}{3pt}
\begin{tabular}{lllcccccc}
\toprule
\rev{Cell} & \rev{Checkpoint} & \rev{Generation} & \rev{HOTA$\uparrow$} & \rev{DetA$\uparrow$} & \rev{AssA$\uparrow$} & \rev{IDF1$\uparrow$} & \multicolumn{2}{c}{\rev{AssA, same boxes}} \\
 & & & & & & & \rev{Kalman/IoU} & \rev{DINOv3 bank} \\
\midrule
\rev{\textbf{A (ours)}} & \rev{prefix-trained} & \rev{chain} & \rev{\textbf{34.8}} & \rev{32.3} & \rev{\textbf{38.3}} & \rev{\textbf{38.3}} & \rev{21.6} & \rev{24.3} \\
\rev{C} & \rev{translation-only} & \rev{chain} & \rev{16.8} & \rev{28.7} & \rev{10.1} & \rev{15.5} & \rev{10.4} & \rev{12.6} \\
\rev{D} & \rev{prefix-trained} & \rev{independent} & \rev{15.9} & \rev{17.4} & \rev{15.1} & \rev{15.7} & \rev{10.7} & \rev{11.2} \\
\rev{B} & \rev{translation-only} & \rev{independent} & \rev{18.2} & \rev{19.7} & \rev{17.7} & \rev{17.3} & \rev{12.7} & \rev{14.3} \\
\midrule
\rev{A, windows re-linked post hoc} & & & \rev{34.7} & \rev{32.4} & \rev{38.0} & \rev{37.7} & & \\
\bottomrule
\end{tabular}
\end{center}
\end{table}

\rev{With the checkpoint fixed, chaining is worth +18.9 HOTA for the prefix-trained
checkpoint (A$-$D, [15.2, 22.9], 24 of 25 videos) and nothing for the translation-only one
(C$-$B, $-$1.2 [$-$3.8, 1.1], 12 of 25). With the generation mode fixed, the checkpoint
does not help independent windows (D$-$B, $-$1.8 [$-$5.1, 0.9], 9 of 25) and is worth +18.3
for chains (A$-$C, [14.9, 22.0], 25 of 25). Neither factor helps on its own; the gain is
their interaction: chaining pays only once the model is trained
to continue the colouring it is handed, which reproduces at full scale the single-video
ablation of Section~\ref{sec:ablations}. Part of each independent cell's deficit is detection: the post-hoc associator,
selected on the 5-person development video, over-merges crowded, identically dressed
people into union boxes (DetA 17.4--19.7). Read by raw colour instead, independent
windows keep their detections (DetA 33.1 in D against 32.3 in A), and the chain still leads
at the same checkpoint by +19.8 HOTA [16.3, 23.6] and +29.9 AssA [24.9, 35.3], in all 25
videos, so the gap is association, not detection.} Applied to the chain's own
windows the same associator is neutral ($-0.1$ HOTA, $-0.4$ AssA combined\rev{; paired
intervals include zero}), so the colours already
carry everything overlap linking can recover. On identical detections (DetA 32.3--32.7), the Kalman/IoU
associator reaches AssA 21.6 and the appearance-and-motion bank 24.3 against 38.3 for the
colours \rev{(+15.3 [11.6, 19.1] and +12.5 [9.0, 16.3], 24 and 23 of 25 videos)}, with
3,079 and 2,408 identity switches against 1,950. Fast dance motion breaks
frame-to-frame geometry and near-identical costumes defeat appearance, exactly where the
generated colour stays stable. \rev{In the independent cells, where colour
carries identity only within a window, the lead over both associators shrinks to +3.1 to
+4.4 AssA but stays positive (all lower bounds $\geq$1.7). In C it does not: chained without
continuation training, the colours are no better than Kalman/IoU ($-$0.5 AssA [$-$1.7, 0.4])
and worse than the bank ($-$2.7 [$-$5.4, $-$1.0]) on the same boxes.} A stronger post-hoc
\rev{window linker} might narrow the A--B gap\rev{, not the trained chain's lead on the same
boxes}.

\subsection{Long-gap re-acquisition}
\label{sec:reacq}

The hardest identity problem is re-acquisition: a person occluded or out of frame for
dozens of frames must return with the same identity. We mined all ground-truth gap events
($\geq$9 absent frames, then reappearance) across the 25 validation videos: 383 events, 360
occlusions and 23 exits (protocol in Appendix~\ref{app:reacq}). For each event we ask
whether the person was tracked before the gap and re-detected after it and, if so,
whether the majority identity matched. Table~\ref{tab:reacq} reports the conditional rate
$k/n$ over such events with its Wilson 95\% interval and the unconditional rate over all
383, because the conditional denominator differs between systems (205 events for the chain,
154 for the independent-window arm), together with the two classical associators run on
the chain's own boxes over the same events.

\begin{table}[t]
\caption{Re-acquisition after ground-truth gaps (383 events, validation set). Conditional
$k/n$ counts events tracked before and re-detected after the gap on the same boxes that
TrackEval scores; the $>$150 bin is included in ``all''. The 49-frame window is the model's
temporal context. The two associators run on arm A's boxes; ``raw colour'' reads arm B's
independently generated windows without any linking.}
\label{tab:reacq}
\begin{center}
\scriptsize
\setlength{\tabcolsep}{3pt}
\begin{tabular}{lccccc}
\toprule
Gap length (frames) & 9--24 & 25--48 & 49--150 & $>$150 & all \\
\midrule
GT gap events & 234 & 88 & 53 & 8 & 383 \\
A: chain (ours), $k/n$ & 60/127 & 19/50 & 7/26 & 0/2 & 86/205 \\
\quad rate (95\% CI) & 0.47 (0.39--0.56) & 0.38 (0.26--0.52) & 0.27 (0.14--0.46) & 0/2 & 0.42 (0.35--0.49) \\
\quad unconditional & 0.26 & 0.22 & 0.13 & 0.00 & 0.22 \\
\quad A boxes, Kalman/IoU, $k/n$ & 33/127 & 12/50 & 2/26 & 0/2 & 47/205 \\
\quad A boxes, appearance$+$motion bank, $k/n$ & 39/127 & 10/50 & 1/26 & 0/2 & 50/205 \\
B: independent $+$ assoc., $k/n$ & 40/96 & 10/38 & 3/19 & 0/1 & 53/154 \\
\quad rate (95\% CI) & 0.42 (0.32--0.52) & 0.26 (0.15--0.42) & 0.16 (0.06--0.38) & 0/1 & 0.34 (0.27--0.42) \\
\quad unconditional & 0.17 & 0.11 & 0.06 & 0.00 & 0.14 \\
\quad B raw colour (no linking), $k/n$ & 21/161 & 5/64 & 5/29 & 0/1 & 31/255 \\
\bottomrule

\end{tabular}
\end{center}
\end{table}

Four observations follow (a qualitative example is in Figure~\ref{fig:reacq},
Appendix~\ref{app:reacq}). First, for gaps shorter than the temporal context (9--48 frames),
the chained model re-binds 47\% and 38\% of
the events it holds, the capability that motivated using a video model. Second, on
identical boxes the colours beat both classical associators: 86 against 47 (Kalman/IoU) and
50 (appearance bank) of the same 205 events; on the paired events the colours win \rev{44} to 5
and \rev{42} to 6 (exact McNemar $p<0.001$). Third, the chain leads the independent-window arm
(22\% against 14\% of all 383 events) but the paired difference is not significant: on the
\rev{118} events both hold, the chain wins 20 and the independent arm 10 ($p=0.10$). Fourth, gaps
longer than the context (49--150 frames) still re-acquire at 27\% (7 of 26), and two
controls say why. Across \emph{independently} generated windows the raw palette colour
agrees before and after the gap in 12\% of events (31 of 255) against a chance level of 5\%
(one over the number of colours in use), a weak colour preference rather than
re-identification; in the chain the pre-gap colour was painted in at least one gap frame for
all 86 re-acquired events, including all 7 beyond the context. \rev{This carry-over is
necessary (in each of the other two seeds, all but one recovery shows it) but not
sufficient: statistics pinned before the run over all 205
conditional events (Appendix~\ref{app:mech}) show that the colour also survives the gap in
110 of the 119 failures. What tends to separate the two outcomes is where the colour stays. In the median recovery it
covers the whole gap (median fraction of gap frames 1.00), is about person-sized (1.11$\times$
the last visible box) and stays near the hidden person's interpolated position (median IoU
0.35). In failures it covers less of the gap (0.90) and drifts away (IoU 0.19). Of the 38
recoveries whose gap spans a window boundary, 37 carry the colour in every intervening
prefix. Of the seven beyond the context, four carry a well-located colour (IoU 0.41--0.64)
through every prefix, two carry a poorly located one (IoU 0.20 and 0.25), and one loses it
in a prefix and recovers anyway (Table~\ref{tab:beyond}). The state-in-pixels mechanism is therefore real
but incomplete: the colour must persist \emph{and} stay attached to its person, and an
identity the model stops painting is lost.}

\begin{table}[t]
\centering
\scriptsize
\setlength{\tabcolsep}{3pt}
\caption{SportsMOT validation with the DanceTrack recipe re-run from the stock backbone,
TrackEval. Left: all 45 validation videos (combined). Right: unweighted mean over the 37
videos that took no part in checkpoint selection (a per-video mean, not a combined score).
Re-acq.: conditional re-acquisitions $k/n$ over all 45 videos under the protocol of
Section~\ref{sec:reacq}.
The zero-shot rows apply the DanceTrack checkpoint with no SportsMOT training to the 8
selection videos only.}
\label{tab:sportsmot}
\begin{tabular}{lccccccccc}
\toprule
& \multicolumn{4}{c}{all val videos (TrackEval combined)} & \multicolumn{4}{c}{held-out mean} & \\
\cmidrule(lr){2-5}\cmidrule(lr){6-9}
association & HOTA & DetA & AssA & IDF1 & HOTA & DetA & AssA & IDF1 & re-acq \\
\midrule
\multicolumn{10}{l}{\textit{Zero-shot (DanceTrack checkpoint)} (8 selection videos)} \\
generated colour & 17.0 & 14.0 & 20.8 & 17.8 & -- & -- & -- & -- & 4/22 \\
Kalman/IoU on the same boxes & 13.3 & 14.1 & 13.0 & 12.7 & -- & -- & -- & -- & 0/22 \\
DINOv3 bank on the same boxes & 14.7 & 14.1 & 15.8 & 14.6 & -- & -- & -- & -- & 5/22 \\
\midrule
\multicolumn{10}{l}{\textit{Fine-tuned on SportsMOT train} (45 videos; 37 held out)} \\
generated colour & 26.9 & 26.0 & 28.2 & 31.9 & 26.9 & 26.3 & 28.4 & 32.2 & 18/159 \\
Kalman/IoU on the same boxes & 20.9 & 26.2 & 17.1 & 22.3 & 21.1 & 26.5 & 17.7 & 22.9 & 0/159 \\
DINOv3 bank on the same boxes & 23.0 & 26.1 & 20.7 & 25.9 & 23.1 & 26.4 & 21.3 & 26.6 & 16/159 \\
\bottomrule

\end{tabular}
\end{table}

\subsection{Generalisation to SportsMOT}
\label{sec:sportsmot}
To test whether the recipe transfers, and not only the DanceTrack model, we repeat it end to
end on SportsMOT \citep{cui2023sportsmot} (45 train / 45 validation videos of basketball,
football and volleyball; validation only, no test submission), re-tuning nothing: the same
SAM~3 labelling of the train split (97.9\% of boxes matched), a fresh LoRA on the stock
backbone with the DanceTrack rank, learning rate and step budget, checkpoints selected by the
same foreground-IoU rule on a fixed 8-video subset, and the chain, decoder and evaluation
unchanged. Table~\ref{tab:sportsmot} reports all 45 validation videos, the 37 outside the
selection subset, and the DanceTrack checkpoint applied zero-shot. Training on the domain
matters: on the eight selection videos mean per-video HOTA rises from 17.3 zero-shot to
28.0, ahead on every video, and the held-out mean (26.9) shows no selection inflation. The
ordering on identical boxes replicates: AssA 28.2 for the colours against 20.7 for the
appearance bank and 17.1 for Kalman/IoU (2,510 identity switches against 3,334 and 5,684).
Long-gap re-acquisition does not carry over, and the reason is instructive: 369 of the 374
SportsMOT gap events are players leaving the frame, not occlusions. Inside the context the
colours still re-bind 15 of the 53 events they hold (chance 0.10; bank 11, Kalman/IoU 0);
beyond it they re-acquire 3 of 106, below chance (0.03 against 0.09), and across all re-acquired events the
pre-gap colour was painted during the gap only half the time. A player who leaves the frame
cannot be kept alive in pixels: the limit Section~\ref{sec:reacq} names, measured where it
dominates.

\subsection{Ablations and analysis}
\label{sec:ablations}

\textbf{Continuation training is necessary.} Single-video ablations use the development
video ($n{=}1$; checkpoints in Appendix~\ref{app:params}). With the same chain at inference
but no prefix training, its full-video IDF1 is 0.172, \emph{below} independent windows,
because the model overrides the imposed colouring mid-window and errors compound. Prefix
training lifts it to 0.466\rev{; cell C confirms this on all 25 videos (Section~\ref{sec:ab})}.
The cost is unconditioned quality:
single-window foreground IoU drops from 0.605 to 0.589, and \rev{in development runs}
independent windows from the prefix-trained checkpoint stitched to IDF1 0.423 against 0.466
from the translation-only checkpoint, which is why arm~B uses the latter
\rev{(the gap is not significant on all 25 videos)}.

\textbf{Window length and chain length.} Generating 121-frame clips from the 49-frame
translation-only checkpoint instead of chaining keeps mask quality but degrades identity
sharply, so we chain at the trained length. Along the chain the loss is mild: full-video
IDF1 retains \rev{about} 85\% of the single-window value and the number of
distinct colours per window does not grow (Appendix~\ref{app:chain}).

\textbf{\rev{Dense scenes fail even on training clips.}} \rev{Failure concentrates in
crowded videos (Figure~\ref{fig:density}). The natural reading is latent-token
capacity, because the chain paints only 10--12 regions per frame on the densest validation
videos. Two measurements contradict that reading (Appendix~\ref{app:dense}). First, a
pre-registered probe renders 12 \emph{training} clips as single windows
with the translation-only checkpoint: on crowds of 23--37 people it paints 16--24 regions per
frame (up to 28), so a $16\times16$-token frame is not capped at 10--12. Yet mask recall at
IoU 0.5 there is only 0.09--0.30, against 0.49--0.94 on sparse training clips, so crowds fail
even on data the model was fit on and training coverage alone does not explain it. Second, a
post-hoc count on validation shows the 10--12 figure is not a fixed budget: on the three
videos averaging $\geq$14 people per frame, independently generated windows paint 14--15
regions per frame with either checkpoint, and chaining lowers the count (12--14 with the
translation-only checkpoint, 10--12 with the prefix-trained one). An exploratory breakdown
of one frame per training clip finds 5--7 regions merging neighbours on the crowded clips
against 0--1 on sparse ones, and 84--97\% of the dancers in those crowds are narrower than one 32-pixel latent token. This
fits a resolution account, as does crowd separation at $768^2$ (Figure~\ref{fig:dense},
where box recall falls because the reference alignment warps). But crowd size and person size vary together in these clips, so
whether per-token resolution, capacity or optimisation is the cause remains unresolved.}

\section{Limitations}
\label{sec:limitations}

Detection \rev{(test DetA 37.6, lowest in dense scenes)} and throughput (about 60 s per 49-frame window on
one A100, 0.7 new frames per second at stride 40 and about \rev{0.4} fps end to end on the test
run, against 20--30 fps for tracking-by-detection) are the two deficits. \rev{Main-table
numbers use one sampling seed. With two further seeds on all 25 validation videos the three
give HOTA 34.8 (reported), 37.1 and 37.8 (sd 1.6), so the reported seed is the lowest, though single videos move
more (median per-video range 8.4 points). Re-acquisition (86/205, 85/226, 85/221), the
carry-over statistics and the lead over both associators on identical boxes (12.5--18.7
AssA) hold in every seed (Appendix~\ref{app:seeds}).} The validation split served as the model-selection and
development set, so its numbers carry selection bias that the test row does not. The
palette bounds cumulative identities at 47 per video and colour reuse is unguarded. Two
domains are covered, choreographed dance and team sports, the second on validation only;
transfer to appearance-diverse pedestrian benchmarks such as MOT17 remains untested. Continuation training teacher-forces ground-truth prefixes, leaving
a train/test gap to self-generated prefixes\rev{, which the decode, correct, re-encode step
narrows but does not remove. Per-window IDF1 falls from 0.54 in windows 1--5 to 0.49 in
windows 16--25 in the reported seed but rises in the other two (Appendix~\ref{app:chain}),
so we see no consistent erosion along the chain; training on self-generated prefixes is
untested}.

\section{Conclusion}

Identity can live in pixels. A video generator, lightly fine-tuned to paint people with
persistent colours and to \emph{continue} a colouring it is handed, tracks full-length
videos on the official DanceTrack server with no separately trained detector, no motion
model, and no association step, with association quality above the 2020--22 benchmark
suite on the benchmark built to break association; modern specialists score higher, and
none share this error profile. Controlled comparisons locate the identity in
generation-time state \rev{that the model is trained to continue: chaining and
continuation training help only together, roughly doubling HOTA;} post-hoc association adds
nothing to the chain's own windows\rev{;} classical associators
on the same boxes fall well short of the colours on two datasets\rev{;} and the generator keeps
occluded identities alive in pixels inside and beyond its context window. We offer this as
task-level evidence for the position of \citet{gabeur2026imagegen} that strong generative
models are generalist vision learners, and that ``generation as the interface'' is a
viable, measurable path for video perception. \rev{It is a proof of concept with two
deficits, crowd detection (cause unresolved) and throughput; whether either moves with the
next generation of video models is untested.}

\subsection*{AI use statement}

In this work, we used generative AI tools to implement methods, to propose and refine
hypotheses, to clean and reformat datasets, to support data analysis, and to interpret
results; the method itself is the authors' own. Concretely, an LLM coding assistant wrote,
modified, and executed the experiment, training, and evaluation scripts under author
direction; proposed candidate diagnoses and control experiments that the authors accepted
or modified; and produced intermediate analyses of logged outputs. We have not used
generative AI tools to design the method, to generate synthetic datasets (silhouette
training labels were produced by a non-generative promptable segmentation model on
ground-truth boxes, Section~\ref{sec:iclora}, rather than annotated by hand), to develop
theoretical models, to formulate mathematical claims, or to translate text; assisting with
proofs is not applicable to this work. Additionally, we used AI tools to create figures,
create and edit code, draft and edit parts of this manuscript, search for information,
format references, and edit for readability.

We have reviewed all AI-assisted work. Every quantitative result reported here originates
from a logged experimental run rather than from model-generated text: the authors fixed
all experimental parameters in a pinned log before each run, and each number in the paper
was traced back to its run record. LLM-written code is covered by an invariant test suite
(including the codec round-trip test of Section~\ref{sec:codec}) run before every
experiment. All baseline values were verified by the authors against the cited primary
sources rather than reproduced from model memory, AI-drafted text was checked by the
authors against the experiment log and revised, and each qualitative example was verified
against the underlying data before inclusion. Research direction, hypotheses accepted for
testing, decisions at experimental gates, compute expenditure, and the final claims were
determined by the authors. We take responsibility for the final content of this work,
including text, claims, or artifacts produced with the aid of generative AI.

\subsection*{Ethics statement}
This work uses the public DanceTrack and SportsMOT benchmarks under their non-commercial
research licences; their videos are Internet-sourced footage of identifiable people
collected by the dataset authors. We add no new personal data and do not redistribute
either dataset. For reference, our figures show frames from DanceTrack validation videos,
and the project page (\url{https://sonam525.github.io/Generator-as-the-tracker/}) shows one full validation video and a short excerpt
of another; all are credited to the dataset authors \citep{sun2022dancetrack}. Released ID-map outputs and checkpoints are for non-commercial research only. No
human-subjects study was conducted.

\subsection*{Reproducibility statement}
All experiments were logged in a pinned experiment record (config hashes, seeds, costs,
and gate verdicts per run). The code, ID-map codec, training and inference configurations,
evaluation scripts and run parameters are at \url{https://github.com/Sonam525/Generator-as-the-tracker}; the LoRA checkpoints
are at \url{https://huggingface.co/Sonam5/Generator-as-the-tracker}. Hyperparameters, prompts, decode rules and the
TrackEval invocation are in Appendix~\ref{app:params}; per-video results, the
re-acquisition protocol, the post-hoc associator, the dense-scene probe and the compute
ledger are in Appendices~\ref{app:pervideo}--\ref{app:compute}. The fine-tune uses a single
80GB GPU. The DanceTrack test result is a single frozen submission to the official server,
with the submission archive and per-sequence outputs included in the repository.

\bibliography{references}
\bibliographystyle{iclr2027_conference}

\appendix

\section{Pinned hyperparameters and additional details}
\label{app:params}
\textbf{Model.} LTX-2.3 dev checkpoint (22B joint audio-video transformer, 48 dual-stream
blocks); Gemma-3 12B text encoder used only to precompute one fixed prompt embedding.
\textbf{LoRA.} Rank 32, $\alpha$ 32, video-stream attention (QKV, output) and feed-forward
projections only; audio stream excluded; 163.6M trainable parameters.
\textbf{Training.} Phase A (translation): 3,000 steps, lr $2\times10^{-4}$, bf16, gradient
checkpointing, flow matching with shifted-logit-normal timestep sampling, batch 1, 777
preprocessed clips ($512^2\times49$), training seed 0; checkpoint at step 1,500 selected by
validation foreground IoU over checkpoints every 500 steps (0.53, 0.56, 0.605, 0.49, 0.58,
0.55). Phase B (continuation): resume from step 1,500 to 3,000 with gradient accumulation 4
and the prefix condition (2 latent frames, $p=0.5$); checkpoint at step 2,500 selected the
same way (foreground IoU 0.589). Foreground IoU is the IoU of the non-black region against
the union of ground-truth silhouettes.
\textbf{Silhouette labels.} SAM~3 prompted with each frame's ground-truth boxes as positive
box exemplars, with the returned masks matched back to those boxes by IoU (97\% of boxes;
filled-box fallback for the rest).
\textbf{Inference.} 49-frame windows, stride 40 (9-frame overlap = prefix length); 20
steps; classifier-free guidance 4.0; spatio-temporal guidance 1.0 on block 29; sampling
seed 0; $512^2$; prompt ``segmentation ID map: paint each person a distinct flat colour on
a black background''; negative prompt ``worst quality, inconsistent motion, blurry,
jittery, distorted''.
\textbf{Decode.} $K{=}48$ palette, index 0 = background; nearest palette colour in CIELAB
for every pixel, so near-black snaps to background; per-colour connected components;
minimum area 150 px; per-frame same-colour union.
\textbf{Windows to tracks.} Overlap ownership split at midpoint; colour index = track id.
\textbf{Association baselines on our own boxes.} The Kalman/IoU baseline is a SORT-style
tracker \citep{bewley2016sort} run on arm A's per-frame boxes with the colour ids dropped.
The appearance-and-motion baseline re-associates the same boxes with a DINOv3 identity bank,
momentum-updated while a track is visible and frozen under occlusion, the bank of our
image-model pilots run with its pinned parameters.
\textbf{Evaluation.} TrackEval with HOTA, CLEAR and Identity metrics, DO\_PREPROC False,
predictions rescaled from $512^2$ to native resolution; motmetrics for IDF1 in the
single-video diagnostics.
\textbf{Checkpoint per experiment.} Arm A, the test submission, the continuation
ablation and the 768$^2$ probe use the phase-B step-2,500 checkpoint \rev{(prefix-trained)}; arm B, the
121-frame test, the oracle decomposition and the 17-window independent numbers use the
phase-A step-1,500 checkpoint \rev{(translation-only)}. The SportsMOT rows of Table~\ref{tab:sportsmot} use the
SportsMOT phase-B step-1,500 checkpoint, trained from the stock backbone with a fresh LoRA
(phase A: best validation foreground IoU 0.555 at step 750 of 3,000, selected by the same
rule; phase B: 0.542 at step 1,500); the zero-shot rows use the DanceTrack phase-B
step-2,500 checkpoint with no SportsMOT training. One validation pass run in a single
process scores \rev{the SportsMOT phase-A step-750 checkpoint and that DanceTrack checkpoint}
on the 8 selection windows at foreground IoU 0.555
and 0.485 respectively, the SportsMOT checkpoint ahead on seven of the eight.

\section{Per-video validation results}
\label{app:pervideo}
\begin{figure}[h]
\centering
\includegraphics[width=0.9\linewidth]{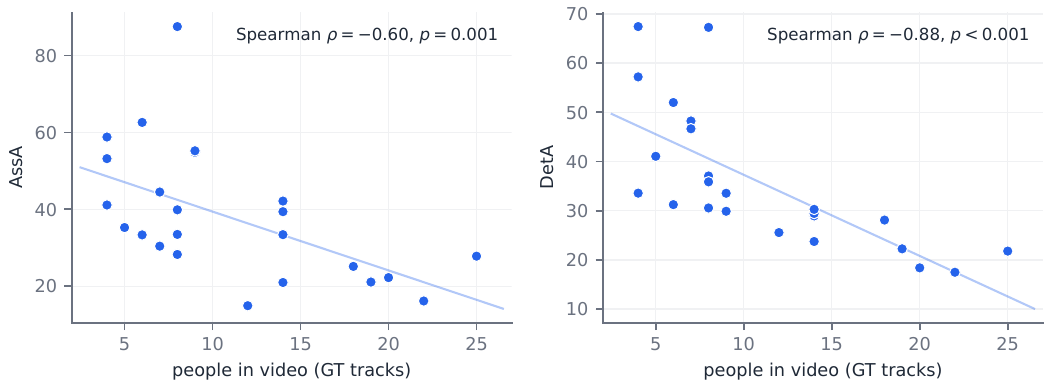}
\caption{\textbf{Association and detection versus scene density} on the 25 validation
videos (chained system). Both fall with the number of people; detection falls faster.}
\label{fig:density}
\end{figure}

Table~\ref{tab:pervideo} lists per-video HOTA components for arms A and B with the number
of ground-truth people and predicted track ids; Figure~\ref{fig:density} plots association
and detection against the number of people. Arm A produces 906 predicted ids for 273
ground-truth identities (3.3 per identity); arm B produces 5,552 (20 per identity). Colour
reuse is visible in the boxes as well: 32\% of arm A's validation boxes span two or more
same-colour components, and the 35 test videos use 24 to 47 of the 47 available colours.
Across the 25 videos, DetA correlates with the number of ground-truth people at Spearman
$\rho = -0.88$ and AssA at $\rho = -0.60$: the 15 videos with at most 9 people average AssA
45.8 and DetA 43.2, and the 9 videos with 14 or more average 27.6 and 24.5.

\begin{table}[h]
\caption{Per-video validation results. People = ground-truth track count; ids = predicted
track ids.}
\label{tab:pervideo}
\begin{center}
\scriptsize
\begin{tabular}{lcc|ccccc|ccccc}
\toprule
 & & & \multicolumn{5}{c|}{A: chain} & \multicolumn{5}{c}{B: independent + assoc.} \\
video & people & frames & HOTA & DetA & AssA & IDF1 & ids & HOTA & DetA & AssA & IDF1 & ids \\
\midrule
0004 & 4 & 1203 & 37.1 & 33.6 & 41.1 & 38.0 & 32 & 20.7 & 23.3 & 19.4 & 26.7 & 226 \\
0005 & 4 & 1203 & 55.1 & 57.2 & 53.2 & 63.1 & 32 & 22.5 & 26.2 & 19.6 & 19.2 & 191 \\
0007 & 8 & 1203 & 32.3 & 37.0 & 28.3 & 42.9 & 40 & 14.4 & 20.6 & 10.3 & 17.1 & 161 \\
0010 & 6 & 1203 & 57.0 & 52.0 & 62.6 & 68.1 & 28 & 43.3 & 41.2 & 45.7 & 49.9 & 185 \\
0014 & 12 & 1203 & 19.4 & 25.6 & 14.9 & 22.9 & 43 & 11.9 & 19.1 & 7.6 & 12.5 & 276 \\
0018 & 8 & 503 & 76.7 & 67.3 & 87.6 & 86.0 & 25 & 51.2 & 38.5 & 68.6 & 60.9 & 96 \\
0019 & 7 & 2402 & 38.3 & 48.2 & 30.4 & 49.1 & 36 & 15.1 & 24.8 & 9.2 & 13.9 & 510 \\
0025 & 9 & 803 & 39.5 & 29.9 & 54.8 & 40.5 & 20 & 28.1 & 33.3 & 24.1 & 31.6 & 102 \\
0026 & 19 & 302 & 21.0 & 22.3 & 21.1 & 22.0 & 45 & 11.3 & 14.8 & 9.6 & 10.9 & 183 \\
0030 & 6 & 1263 & 32.2 & 31.2 & 33.3 & 35.7 & 22 & 18.2 & 29.0 & 11.6 & 17.4 & 104 \\
0034 & 14 & 923 & 30.9 & 29.0 & 33.4 & 32.5 & 45 & 12.1 & 17.0 & 8.9 & 13.0 & 339 \\
0035 & 8 & 703 & 38.3 & 37.1 & 39.9 & 50.1 & 34 & 23.7 & 28.5 & 19.8 & 26.0 & 219 \\
0041 & 22 & 1003 & 16.7 & 17.5 & 16.1 & 18.1 & 43 & 7.4 & 10.6 & 5.4 & 8.1 & 315 \\
0043 & 14 & 183 & 34.4 & 29.4 & 42.1 & 40.1 & 37 & 36.1 & 28.6 & 46.7 & 42.5 & 76 \\
0047 & 8 & 1203 & 28.9 & 30.6 & 28.2 & 31.4 & 38 & 10.0 & 16.3 & 6.3 & 12.5 & 217 \\
0058 & 7 & 1601 & 45.5 & 46.7 & 44.5 & 51.2 & 39 & 16.3 & 22.3 & 12.0 & 15.4 & 240 \\
0063 & 8 & 1000 & 34.5 & 35.9 & 33.5 & 40.7 & 45 & 8.6 & 10.8 & 7.4 & 5.6 & 458 \\
0065 & 5 & 702 & 38.0 & 41.1 & 35.2 & 46.6 & 30 & 21.4 & 28.5 & 16.3 & 22.7 & 155 \\
0073 & 14 & 703 & 34.1 & 30.3 & 39.4 & 37.1 & 46 & 20.2 & 17.9 & 23.0 & 21.9 & 192 \\
0077 & 9 & 1203 & 42.6 & 33.5 & 55.2 & 47.9 & 25 & 14.5 & 15.0 & 14.1 & 15.7 & 132 \\
0079 & 18 & 1202 & 26.4 & 28.1 & 25.1 & 30.7 & 39 & 13.6 & 17.7 & 10.7 & 12.5 & 181 \\
0081 & 20 & 984 & 19.9 & 18.4 & 22.2 & 22.2 & 44 & 8.2 & 12.7 & 5.5 & 11.0 & 278 \\
0090 & 14 & 1004 & 22.0 & 23.7 & 20.9 & 24.9 & 46 & 12.8 & 13.8 & 12.2 & 12.2 & 343 \\
0094 & 25 & 603 & 24.1 & 21.8 & 27.8 & 23.2 & 46 & 8.2 & 9.4 & 7.8 & 7.6 & 222 \\
0097 & 4 & 1203 & 62.9 & 67.4 & 58.8 & 78.5 & 26 & 32.0 & 41.9 & 24.6 & 34.6 & 151 \\
\midrule
all & -- & -- & 34.8 & 32.3 & 38.3 & 38.3 & 906 & 18.2 & 19.7 & 17.7 & 17.3 & 5552 \\
\bottomrule

\end{tabular}
\end{center}
\end{table}

\section{Re-acquisition protocol}
\label{app:reacq}
\begin{figure}[h]
\centering
\includegraphics[width=0.7\linewidth]{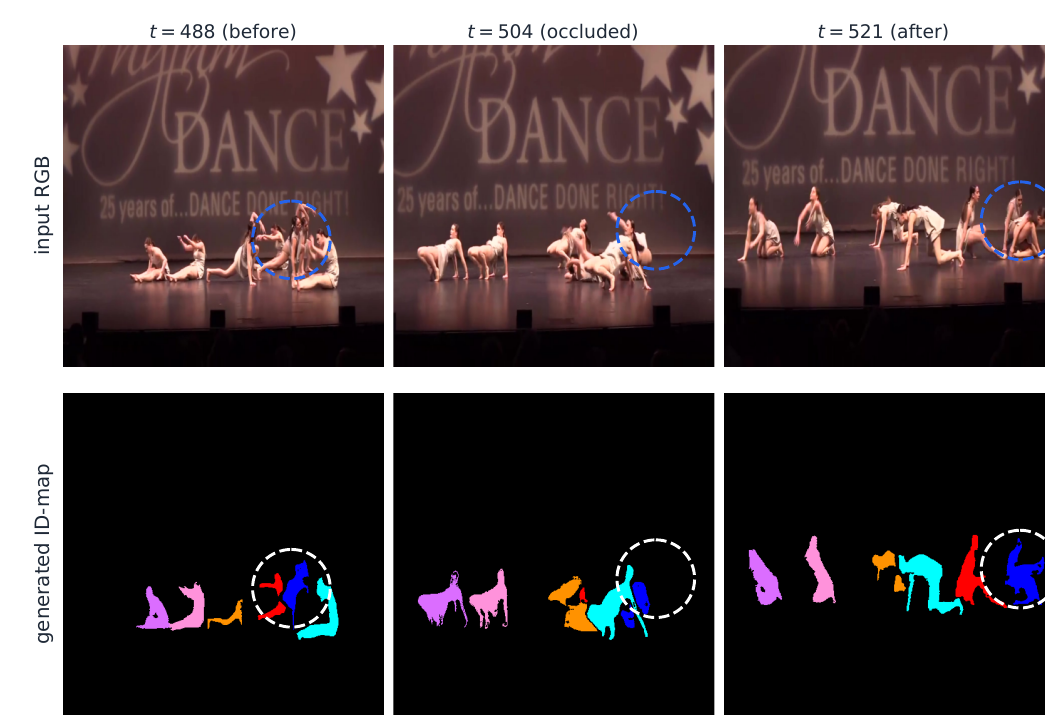}
\caption{\textbf{Re-acquisition through occlusion.} A dancer (dashed circle) is absent
from the ground truth for 27 frames and returns painted the \emph{same blue}; the
generator, not an association module, re-binds the identity. The small same-colour
fragment that persists during the occlusion (middle) is the carry-over: in every
re-acquired event \rev{of the reported seed, and in all but one in each of the other two seeds,}
the colour was painted somewhere during the gap\rev{. It also
survives in 110 of 119 failures, so carry-over is necessary but not sufficient
(Appendix~\ref{app:mech})}.}
\label{fig:reacq}
\end{figure}

Gap events are mined from the ground truth of each validation video: for every track and
every pair of consecutive annotated frames $f_{i-1}, f_i$ with $f_i - f_{i-1} - 1 \geq 9$,
one event with gap $f_i - f_{i-1} - 1$; it is an ``exit'' if the last pre-gap box touches
the frame border (2\% margin), else an ``occlusion''. For a tracker output, the pre-gap
identity is the majority predicted id over the nearest 15 visible ground-truth frames
before the gap among frames where a predicted box overlaps the ground-truth box at IoU
$\geq 0.3$; the post-gap identity is defined symmetrically. \emph{Tracked before} and
\emph{re-detected} require at least 3 matched frames on the respective side; an event is
conditional if both hold, and re-acquired if the two majority ids coincide. Bins are by
gap length; the 8 events longer than 150 frames (none re-acquired by either arm) appear
only in the ``all'' column of Table~\ref{tab:reacq}. All parameters were fixed before the
study. Figure~\ref{fig:reacq} shows one event.

\rev{\section{What carries identity through a gap}
\label{app:mech}
For every conditional gap event of the chain (raw-colour link) we computed, with definitions
pinned before the run: the fraction of gap frames in which the pre-gap colour is present; its
number of runs and longest run; its mean box area relative to the person's last visible
ground-truth box; the IoU of its box with the person's linearly interpolated ground-truth
box; the fraction of its frames in which it overlaps another visible person at IoU
$\geq$0.5; its connected-component count; and, for every window whose start falls inside the
gap, the number of the 9 conditioning-prefix frames that contain it.
Table~\ref{tab:mech} gives medians, split by outcome; Table~\ref{tab:beyond} lists the seven
recoveries beyond the 49-frame context. One further quantity was added after these
statistics were computed and is exploratory: in the frames where the carried colour lies on
another visible person, how often that person also has a box of its own in another colour
(IoU $\geq$0.3). The median is 0.04 of those frames for recoveries (mean 0.31, 73 events) and
0.25 for failures (mean 0.34, 77 events); in seeds 1 and 2 (Appendix~\ref{app:seeds}) the
medians are 0.00--0.09. During the overlap the occluder is therefore mostly painted in the
hidden person's colour rather than its own, so carrying one identity can cost the occluder
its own in those frames.}

\begin{table}[h]
\caption{\rev{Carry-over statistics, chain (arm A), all 205 conditional gap events
(medians unless counted). Area, IoU, on-another-person and component medians are over the
events with the colour in at least one gap frame (86 and 110), where they are defined.}}
\label{tab:mech}
\begin{center}
\scriptsize
\begin{tabular}{lcc}
\toprule
 & \rev{Re-acquired ($n{=}86$)} & \rev{Not re-acquired ($n{=}119$)} \\
\midrule
\rev{colour present in $\geq$1 gap frame} & \rev{86} & \rev{110} \\
\rev{fraction of gap frames with the colour} & \rev{1.00} & \rev{0.90} \\
\rev{runs / longest run (frames)} & \rev{1 / 14} & \rev{1 / 10} \\
\rev{gap length (frames)} & \rev{16} & \rev{20} \\
\rev{area relative to last visible box} & \rev{1.11$\times$} & \rev{0.93$\times$} \\
\rev{IoU with interpolated position} & \rev{0.35} & \rev{0.19} \\
\rev{fraction of frames on another visible person} & \rev{0.62} & \rev{0.32} \\
\rev{connected components} & \rev{1.24} & \rev{1.21} \\
\rev{gap spans a window start} & \rev{38} & \rev{76} \\
\rev{\quad colour in every intervening prefix} & \rev{37} & \rev{65} \\
\rev{\quad \ldots in all 9 frames of every prefix} & \rev{32} & \rev{40} \\
\rev{\quad colour in no prefix} & \rev{0} & \rev{9} \\
\bottomrule
\end{tabular}
\end{center}
\end{table}

\begin{table}[h]
\caption{\rev{The seven re-acquisitions beyond the 49-frame context (chain). Present =
fraction of gap frames with the pre-gap colour; prefix frames = frames of each intervening
9-frame prefix containing it; IoU = with the interpolated position and area = relative to
the last visible box, each a mean over the gap frames in which the colour is present.}}
\label{tab:beyond}
\begin{center}
\scriptsize
\begin{tabular}{lccccc}
\toprule
\rev{Video / id} & \rev{Gap} & \rev{Present} & \rev{Prefix frames} & \rev{IoU} & \rev{Area} \\
\midrule
\rev{0010 / 3} & \rev{102} & \rev{1.00} & \rev{9, 9} & \rev{0.62} & \rev{1.44$\times$} \\
\rev{0014 / 11} & \rev{71} & \rev{1.00} & \rev{9} & \rev{0.58} & \rev{1.04$\times$} \\
\rev{0073 / 7} & \rev{50} & \rev{1.00} & \rev{9, 9} & \rev{0.41} & \rev{1.79$\times$} \\
\rev{0081 / 19} & \rev{130} & \rev{0.88} & \rev{9, 6, 3} & \rev{0.64} & \rev{1.04$\times$} \\
\rev{0034 / 10} & \rev{83} & \rev{0.75} & \rev{9, 9} & \rev{0.20} & \rev{4.35$\times$} \\
\rev{0014 / 8} & \rev{54} & \rev{0.57} & \rev{2} & \rev{0.25} & \rev{1.52$\times$} \\
\rev{0034 / 12} & \rev{83} & \rev{0.48} & \rev{9, 0} & \rev{0.07} & \rev{2.07$\times$} \\
\bottomrule
\end{tabular}
\end{center}
\end{table}

\rev{\section{Sampling seeds}
\label{app:seeds}
Seeds 1 and 2 repeat the chain of arm A (prefix-trained checkpoint) on all 25 validation
videos; only the sampling seed changes. Table~\ref{tab:seeds} gives the combined scores and
the carry-over statistics of Appendix~\ref{app:mech} per seed, and Table~\ref{tab:seedsvideo}
the per-video results, with the two associators of Section~\ref{sec:ab} re-run on each
seed's own boxes. Chance re-acquisition is 0.076--0.079 in every seed. Failures also carry
the colour in every seed (in 110 of 119, 125 of 141 and 123 of 136 for seeds 0, 1 and 2, the
order of every list below). Beyond the context
(49--150 frames) the seeds re-acquire 7 of 26, 10 of 29 and 7 of 27 events, with the colour in every intervening
prefix for 6, 7 and 7 of them. On each seed's own boxes the colours lead both associators in
AssA (paired per-video bootstrap): Kalman/IoU by +15.3 [11.6, 19.1], +18.7 [14.4, 23.2] and
+18.3 [13.7, 23.4], in 24, 25 and 25 of 25 videos, and the DINOv3 bank by +12.5 [9.0, 16.3],
+14.8 [11.6, 18.2] and +15.5 [11.3, 20.2], in 23, 24 and 25 of 25 videos, with DetA within 0.6.}

\begin{table}[h]
\caption{\rev{Three sampling seeds of the chain, 25 validation videos. Carried = re-acquired
events whose pre-gap colour is painted in at least one gap frame; IoU = median IoU of the
carried colour with the hidden person's interpolated position, re-acquired / not
re-acquired.}}
\label{tab:seeds}
\begin{center}
\scriptsize
\setlength{\tabcolsep}{3pt}
\begin{tabular}{lcccccccc}
\toprule
\rev{Seed} & \rev{HOTA} & \rev{DetA} & \rev{AssA} & \rev{IDF1} & \rev{IDSW} & \rev{Re-acq.\ $k/n$} & \rev{Carried} & \rev{IoU} \\
\midrule
\rev{0 (reported)} & \rev{34.8} & \rev{32.3} & \rev{38.3} & \rev{38.3} & \rev{1,950} & \rev{86/205} & \rev{86/86} & \rev{0.35 / 0.19} \\
\rev{1} & \rev{37.1} & \rev{32.3} & \rev{43.5} & \rev{40.1} & \rev{1,755} & \rev{85/226} & \rev{84/85} & \rev{0.37 / 0.14} \\
\rev{2} & \rev{37.8} & \rev{33.5} & \rev{43.7} & \rev{40.9} & \rev{1,908} & \rev{85/221} & \rev{84/85} & \rev{0.39 / 0.13} \\
\midrule
\rev{mean $\pm$ sd} & \rev{36.5 $\pm$ 1.6} & \rev{32.7 $\pm$ 0.7} & \rev{41.8 $\pm$ 3.0} & \rev{39.8 $\pm$ 1.4} & \rev{1,871 $\pm$ 103} & & & \\
\bottomrule
\end{tabular}
\end{center}
\end{table}

\begin{table}[h]
\caption{\rev{Per-video results for seeds 0/1/2. HOTA and AssA of the colours; AssA of
Kalman/IoU and of the DINOv3 bank re-run on the same seed's boxes. People = ground-truth
identities.}}
\label{tab:seedsvideo}
\begin{center}
\scriptsize
\setlength{\tabcolsep}{2.5pt}
\begin{tabular}{lc|ccc|ccc|ccc|ccc}
\toprule
 & & \multicolumn{3}{c|}{\rev{HOTA, colours}} & \multicolumn{3}{c|}{\rev{AssA, colours}} & \multicolumn{3}{c|}{\rev{AssA, Kalman/IoU}} & \multicolumn{3}{c}{\rev{AssA, DINOv3 bank}} \\
\rev{video} & \rev{people} & 0 & 1 & 2 & 0 & 1 & 2 & 0 & 1 & 2 & 0 & 1 & 2 \\
\midrule
\rev{0004} & \rev{4} & \rev{37.1} & \rev{42.0} & \rev{33.6} & \rev{41.1} & \rev{41.3} & \rev{29.5} & \rev{17.0} & \rev{16.9} & \rev{14.0} & \rev{18.9} & \rev{25.4} & \rev{14.3} \\
\rev{0005} & \rev{4} & \rev{55.1} & \rev{64.5} & \rev{64.2} & \rev{53.2} & \rev{68.0} & \rev{68.9} & \rev{20.5} & \rev{24.5} & \rev{19.4} & \rev{19.4} & \rev{40.8} & \rev{27.9} \\
\rev{0007} & \rev{8} & \rev{32.3} & \rev{33.9} & \rev{40.5} & \rev{28.3} & \rev{49.3} & \rev{41.1} & \rev{18.0} & \rev{20.3} & \rev{17.4} & \rev{18.5} & \rev{22.7} & \rev{18.2} \\
\rev{0010} & \rev{6} & \rev{57.0} & \rev{69.3} & \rev{55.2} & \rev{62.6} & \rev{71.1} & \rev{58.1} & \rev{30.4} & \rev{31.8} & \rev{28.7} & \rev{39.6} & \rev{36.7} & \rev{33.7} \\
\rev{0014} & \rev{12} & \rev{19.4} & \rev{22.5} & \rev{24.0} & \rev{14.9} & \rev{17.4} & \rev{18.9} & \rev{7.0} & \rev{11.2} & \rev{9.2} & \rev{9.1} & \rev{11.2} & \rev{11.0} \\
\rev{0018} & \rev{8} & \rev{76.7} & \rev{54.0} & \rev{70.4} & \rev{87.6} & \rev{73.9} & \rev{85.7} & \rev{66.6} & \rev{57.9} & \rev{71.0} & \rev{80.4} & \rev{64.3} & \rev{81.6} \\
\rev{0019} & \rev{7} & \rev{38.3} & \rev{35.3} & \rev{27.7} & \rev{30.4} & \rev{28.0} & \rev{24.1} & \rev{9.0} & \rev{10.8} & \rev{11.6} & \rev{13.6} & \rev{12.6} & \rev{17.6} \\
\rev{0025} & \rev{9} & \rev{39.5} & \rev{50.5} & \rev{46.3} & \rev{54.8} & \rev{55.2} & \rev{55.0} & \rev{48.4} & \rev{38.9} & \rev{31.3} & \rev{51.7} & \rev{47.2} & \rev{46.4} \\
\rev{0026} & \rev{19} & \rev{21.0} & \rev{19.7} & \rev{15.2} & \rev{21.1} & \rev{18.8} & \rev{12.8} & \rev{13.3} & \rev{14.8} & \rev{8.3} & \rev{14.2} & \rev{11.9} & \rev{6.9} \\
\rev{0030} & \rev{6} & \rev{32.2} & \rev{35.0} & \rev{31.7} & \rev{33.3} & \rev{33.4} & \rev{33.1} & \rev{13.1} & \rev{17.3} & \rev{15.4} & \rev{21.6} & \rev{17.8} & \rev{11.4} \\
\rev{0034} & \rev{14} & \rev{30.9} & \rev{29.3} & \rev{33.8} & \rev{33.4} & \rev{29.4} & \rev{37.4} & \rev{15.8} & \rev{14.4} & \rev{18.9} & \rev{17.4} & \rev{15.4} & \rev{24.3} \\
\rev{0035} & \rev{8} & \rev{38.3} & \rev{46.4} & \rev{45.9} & \rev{39.9} & \rev{52.4} & \rev{46.5} & \rev{13.0} & \rev{20.4} & \rev{16.8} & \rev{20.1} & \rev{19.5} & \rev{21.0} \\
\rev{0041} & \rev{22} & \rev{16.7} & \rev{16.9} & \rev{19.2} & \rev{16.1} & \rev{16.6} & \rev{18.2} & \rev{9.8} & \rev{8.6} & \rev{10.6} & \rev{11.4} & \rev{7.8} & \rev{10.1} \\
\rev{0043} & \rev{14} & \rev{34.4} & \rev{34.0} & \rev{33.3} & \rev{42.1} & \rev{38.6} & \rev{41.4} & \rev{39.6} & \rev{34.6} & \rev{35.7} & \rev{43.9} & \rev{40.0} & \rev{36.7} \\
\rev{0047} & \rev{8} & \rev{28.9} & \rev{40.0} & \rev{22.1} & \rev{28.2} & \rev{45.7} & \rev{26.5} & \rev{29.1} & \rev{29.9} & \rev{13.1} & \rev{29.3} & \rev{31.1} & \rev{15.8} \\
\rev{0058} & \rev{7} & \rev{45.5} & \rev{49.7} & \rev{66.2} & \rev{44.5} & \rev{58.1} & \rev{63.6} & \rev{20.4} & \rev{27.4} & \rev{35.8} & \rev{19.1} & \rev{45.3} & \rev{44.9} \\
\rev{0063} & \rev{8} & \rev{34.5} & \rev{22.7} & \rev{31.1} & \rev{33.5} & \rev{26.2} & \rev{25.8} & \rev{17.3} & \rev{8.8} & \rev{17.7} & \rev{21.9} & \rev{13.1} & \rev{19.4} \\
\rev{0065} & \rev{5} & \rev{38.0} & \rev{40.0} & \rev{43.8} & \rev{35.2} & \rev{35.7} & \rev{53.7} & \rev{10.7} & \rev{14.6} & \rev{20.0} & \rev{10.3} & \rev{13.7} & \rev{13.6} \\
\rev{0073} & \rev{14} & \rev{34.1} & \rev{24.6} & \rev{24.3} & \rev{39.4} & \rev{29.6} & \rev{35.1} & \rev{29.7} & \rev{22.2} & \rev{17.7} & \rev{33.1} & \rev{23.3} & \rev{15.7} \\
\rev{0077} & \rev{9} & \rev{42.6} & \rev{55.7} & \rev{63.5} & \rev{55.2} & \rev{61.5} & \rev{66.8} & \rev{45.9} & \rev{48.0} & \rev{45.2} & \rev{39.3} & \rev{45.0} & \rev{46.7} \\
\rev{0079} & \rev{18} & \rev{26.4} & \rev{23.8} & \rev{28.8} & \rev{25.1} & \rev{29.5} & \rev{32.7} & \rev{20.3} & \rev{25.2} & \rev{30.0} & \rev{21.7} & \rev{25.2} & \rev{29.6} \\
\rev{0081} & \rev{20} & \rev{19.9} & \rev{20.3} & \rev{20.5} & \rev{22.2} & \rev{22.7} & \rev{20.7} & \rev{10.4} & \rev{11.6} & \rev{12.9} & \rev{12.2} & \rev{11.8} & \rev{15.0} \\
\rev{0090} & \rev{14} & \rev{22.0} & \rev{27.3} & \rev{31.0} & \rev{20.9} & \rev{34.5} & \rev{35.5} & \rev{14.3} & \rev{13.3} & \rev{24.4} & \rev{17.3} & \rev{18.4} & \rev{21.9} \\
\rev{0094} & \rev{25} & \rev{24.1} & \rev{29.6} & \rev{20.1} & \rev{27.8} & \rev{39.5} & \rev{22.8} & \rev{19.7} & \rev{27.1} & \rev{18.8} & \rev{22.2} & \rev{26.9} & \rev{22.3} \\
\rev{0097} & \rev{4} & \rev{62.9} & \rev{62.0} & \rev{61.2} & \rev{58.8} & \rev{59.3} & \rev{64.0} & \rev{27.9} & \rev{18.5} & \rev{16.1} & \rev{30.1} & \rev{38.4} & \rev{23.2} \\
\midrule
\rev{combined} & \rev{273} & \rev{34.8} & \rev{37.1} & \rev{37.8} & \rev{38.3} & \rev{43.5} & \rev{43.7} & \rev{21.6} & \rev{23.1} & \rev{23.9} & \rev{24.3} & \rev{27.4} & \rev{27.6} \\
\bottomrule

\end{tabular}
\end{center}
\end{table}

\section{The post-hoc associator (arm B)}
\label{app:stitcher}
Windows are decoded independently. Between consecutive windows, each colour of window $k$
is linked to the colour of window $k{-}1$ with the largest summed box IoU over the 9
overlap frames if that sum exceeds 0.5, many-to-one (several current colours may join one
previous track, which absorbs within-window colour splits); unlinked colours start new
ids, and same-frame boxes of one id are merged by union. The design was selected on the
development video by an oracle decomposition: one-to-one Hungarian linking reached IDF1
0.413, many-to-one 0.466, and an oracle stitcher 0.698 (in-window mean 0.661). Global
geometric gap healing and colour-affinity healing added nothing. On the development video
the associator ties the chain (0.454 against 0.466); at scale it collapses on dense scenes
(Table~\ref{tab:ab}).

\section{Chain length}
\label{app:chain}
\begin{figure}[h]
\centering
\includegraphics[width=0.9\linewidth]{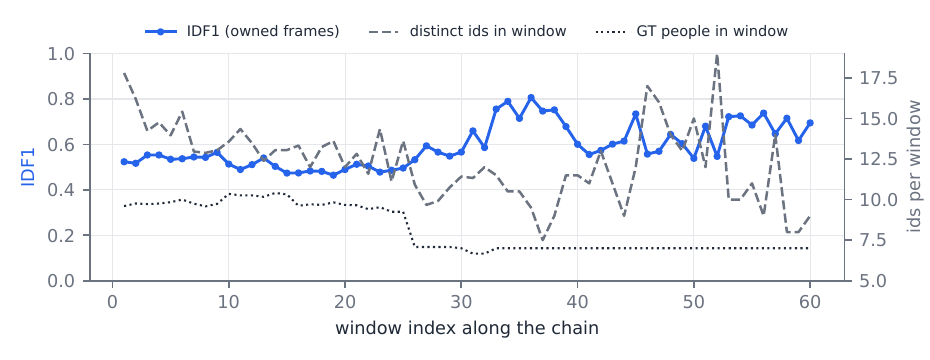}
\caption{\textbf{Identity quality along the chain.} Per-window IDF1 (left axis) and
distinct colours per window against ground-truth people (right axis), mean over validation
videos reaching each window index.}
\label{fig:decay}
\end{figure}

Figure~\ref{fig:decay} plots, for the chained system, IDF1 over each window's owned frames
and the number of distinct colours per window against the window index, averaged over the
validation videos that reach that index (25 at window 1, 21 at window 18, 12 at window 30,
one at window 60). \rev{In the reported seed,} IDF1 drifts from about \rev{0.54} to about 0.49 over the first 25 windows
\rev{(windows 1--5 against 16--25), which amounts to 0.05 over about 1,000 frames; in seeds 1
and 2 it rises instead, from 0.49 and 0.51 to 0.54, so the drift is not systematic. The}
colour count tracks the number of people rather than growing, so fragmentation does
not accumulate along the chain; the rise after window 30 reflects the few long videos
that reach those indices. Full-video IDF1 averages 0.417 per video, \rev{about} 85\% of the 0.488 mean
over 49-frame single windows\rev{; that mean is from the translation-only checkpoint, so the
ratio mixes the checkpoint and chaining effects}.
Generating longer clips instead of chaining is worse: 121-frame clips from the 49-frame
translation-only checkpoint keep mask quality but lose identity, scoring IDF1 0.549 over
frames 1--49 against 0.312 over frames 50--121, with a colour-switch rate $2.3\times$ higher.

\section{Dense-scene probe}
\label{app:dense}
\begin{figure}[h]
\centering
\includegraphics[width=0.9\linewidth]{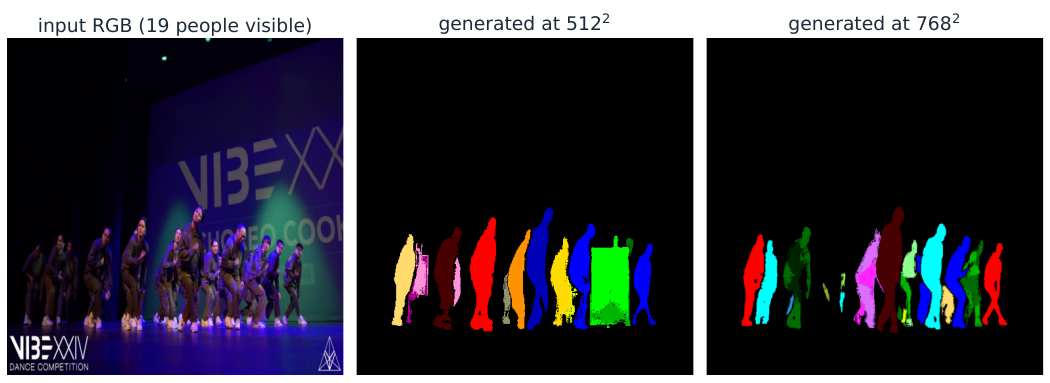}
\caption{\textbf{\rev{Crowd separation at two resolutions}} \rev{(chain, prefix-trained
checkpoint).} At $512^2$ (16$\times$16 latent tokens per frame) the chain paints only
$\sim$12 of 19 people, merges neighbours, and hallucinates a non-person region; at $768^2$,
with the \emph{same} checkpoint, crowd separation improves dramatically, but reference
alignment warps at the unseen resolution, so boxes degrade. \rev{The count of $\sim$12 is
not a token-budget ceiling (Appendix~\ref{app:dense}, Table~\ref{tab:densetrain}).}}
\label{fig:dense}
\end{figure}

\rev{\textbf{Validation diagnosis (chain, prefix-trained checkpoint).}}
Per-frame counts and recall of large ($\geq$80 px) people for the videos used in the
dense-scene diagnosis at $512^2$: dancetrack0041 (16.7 people per frame, 10.1 painted,
merge rate 0.19, recall 0.41), 0081 (19.1 / 11.0 / 0.16 / 0.45), 0094 (19.5 / 12.1 / 0.20 /
0.48), 0026 (13.2 / 12.4 / 0.22 / 0.58), 0014 (10.5 / 7.6 / 0.32 / 0.71), 0018 (5.8 / 5.6 /
0.17 / 0.86), 0097 (3.9 / 4.4 / 0.14 / 1.00). Regenerating dancetrack0094 at $768^2$ with the
same checkpoint separates crowds but loses boxes: 13.1 painted per frame, merge rate 0.20 to
0.11, recall of large people 0.48 to 0.32 (Figure~\ref{fig:dense}). Full-video IDF1 follows
the same axis: \rev{0.31}--0.86 on the \rev{15} videos with 4--9 people against 0.18--\rev{0.40} on the \rev{9} with
14--25\rev{. Read alone, these counts suggest a latent-token-capacity ceiling;
the measurements below contradict that reading.}

\rev{\textbf{Training-clip probe (pre-registered).} We rendered 12 \emph{training} clips
that the translation-only checkpoint was fit on, as single 49-frame windows at $512^2$ with
its validation sampler (seed 0, 20 steps, guidance 4.0). The clips were chosen by a rule fixed
before the run. It takes four clips from each of two density bands ($\geq$20 annotated people
per frame, and 14--19), each clip being the densest window of a training sequence within its band (two
sequences give one clip to each band), and adds four sparse calibration clips
(Table~\ref{tab:densetrain}). The pre-registered reading rule had two
branches: if painted regions stay at or below about 12 while the ground truth has 14--40
people, the ceiling holds on seen data (capacity or fitting); if the model paints close to
the ground-truth count, the validation ceiling is a coverage effect. The outcome matched
neither. The model paints 16--24 regions per frame on the $\geq$20-person clips (59--81\% of
the ground-truth count), so the ceiling does not hold, yet recall is 0.09--0.30, so the
failure is present on seen data. Crowds are in the training set: of the 779 windows of 49 frames
we count in the training annotations, 95 average $\geq$14 people per frame and 65 average $\geq$20, although the 14--19 band is
thinner than in validation (3.7\% of training frames against 11.1\%).}

\begin{table}[h]
\caption{\rev{Density probe on training clips (translation-only checkpoint, single windows,
$512^2$). Painted = decoded regions per frame (minimum area 150 px); recall = Hungarian
instance recall at mask IoU $\geq$0.5. Clip = sequence and first frame of the window. Band =
the pre-registered selection by annotated people per frame; GT / frame counts decoded
ground-truth regions with the same minimum area, which drops small or occluded people.}}
\label{tab:densetrain}
\begin{center}
\scriptsize
\begin{tabular}{llccc}
\toprule
\rev{Band} & \rev{Clip} & \rev{GT / frame} & \rev{Painted / frame (max)} & \rev{Recall} \\
\midrule
\rev{$\geq$20} & \rev{0096 / 1} & \rev{37.4} & \rev{23.8 (28)} & \rev{0.09} \\
 & \rev{0020 / 295} & \rev{32.1} & \rev{19.0 (22)} & \rev{0.17} \\
 & \rev{0083 / 1} & \rev{24.0} & \rev{19.5 (23)} & \rev{0.30} \\
 & \rev{0082 / 1} & \rev{22.9} & \rev{15.7 (18)} & \rev{0.16} \\
\rev{14--19} & \rev{0096 / 393} & \rev{16.0} & \rev{10.1 (12)} & \rev{0.20} \\
 & \rev{0080 / 638} & \rev{14.7} & \rev{12.9 (15)} & \rev{0.31} \\
 & \rev{0086 / 295} & \rev{15.0} & \rev{12.4 (14)} & \rev{0.35} \\
 & \rev{0082 / 442} & \rev{13.9} & \rev{8.5 (11)} & \rev{0.26} \\
\rev{4--9} & \rev{0001 / 1} & \rev{6.0} & \rev{7.4 (9)} & \rev{0.94} \\
 & \rev{0002 / 1} & \rev{7.0} & \rev{8.9 (10)} & \rev{0.52} \\
 & \rev{0006 / 1} & \rev{7.8} & \rev{9.5 (11)} & \rev{0.49} \\
 & \rev{0008 / 1} & \rev{7.0} & \rev{8.3 (10)} & \rev{0.90} \\
\bottomrule
\end{tabular}
\end{center}
\end{table}

\rev{\textbf{Exploratory, post hoc.} (i) We counted painted regions per frame (TrackEval
detection counts under the raw-colour link) on the three validation videos averaging
$\geq$14 people per frame (0094, 0081 and 0041, with 19.5, 19.1 and 16.7 people). Independent windows paint
14.3, 14.7 and 14.9 regions with the translation-only checkpoint and 14.8, 13.6 and 14.8 with
the prefix-trained one. Chains paint fewer: 13.7, 12.1 and 12.3 with the translation-only
checkpoint and 12.1, 11.0 and 10.1 with the prefix-trained one. The 10--12 figure therefore comes
from chained generation, most strongly with the prefix-trained checkpoint, not from the
token budget. (ii) Over the ground-truth frames of
the training clips, the median dancer is 20--23 px wide on the $\geq$20-person clips
(0.62--0.72 of a 32-px latent token; 84--97\% of people narrower than one token) against
32--46 px on the sparse clips; in frame 25 of each clip, 5--7 generated masks merge two
or more people on the crowded clips, against 0--1 on the sparse ones. Visually, the model paints person-shaped regions over the crowd at a
coarser granularity than the dancers. Density and person size are confounded in these
clips, so these numbers do not separate resolution from capacity or optimisation.}

\section{Compute ledger}
\label{app:compute}
A100-hours per stage, from the run records \rev{unless marked}: SAM~3 silhouette labels 8.5; \rev{LTX
feasibility probe 1.2 (the conditioning sanity runs that followed it were not logged)}; Phase A
training with six in-loop validation passes over the 25 validation videos $\approx$26 \rev{and
Phase B with three $\approx$17; development-video chains, the
length test and the overlap-prefix test 3.9}; the two full-scale validation arms 22; the $768^2$ probe 1.5; test-set generation 26.
\rev{The two training figures marked $\approx$ are upper estimates from the logged step time and
an assumed 4 hours per validation pass, because the run records hold no wall-clock totals;
the log notes about 6 A100-hours for the Phase B run, which matches the logged step time
for its optimiser steps alone.}
SportsMOT arm, same recipe from the stock backbone: SAM~3 labels 5.3; zero-shot chain 4.3;
Phase A 3.7; Phase B 7.8; the 45-video validation chain 20.2 (two launches, the first
reaching 41 videos at its 18-hour limit); the single-process validation pass behind the
foreground-IoU comparison 0.5. Three-seed check on five DanceTrack videos 7.
\rev{Factorial cells C and D 18.8 each; seeds 1 and 2 extended to all 25
validation videos 15.8 and 15.4 (the pre-registered estimate for these four runs was about 40);
the training-clip density probe 0.6.} Total $\approx$\rev{224} A100-hours. \rev{Evaluation
(TrackEval, associators, re-acquisition and mechanism statistics) ran on single T4 GPUs:
about 9.1 T4-hours for the runs listed before the factorial cells (5.5 of them for the
association baselines); 5.8 lost to a parameter-handling bug; and 14.9 for the factorial
cells, the two further seeds and the mechanism statistics (cells C and D 2.5 and 2.4; seeds
4.3 and their associators 4.7; mechanism statistics 1.0). In all that is about 30 T4-hours,
plus CPU time.} The image-model pilots that
preceded the video arm ($\approx$\rev{25--30} A100-hours \rev{besides the SAM~3 labels counted
above}, and about \$21 of image-editing API calls) are not included.

\end{document}